\documentclass[letterpaper]{article} 
\usepackage[submission]{aaai23}  
\usepackage{times}  
\usepackage{helvet}  
\usepackage{courier}  
\usepackage[hyphens]{url}  
\usepackage{graphicx} 
\usepackage{natbib}  
\usepackage{caption} 
\usepackage{algorithm}
\usepackage{algorithmic}
\usepackage{multirow}
\usepackage{amsmath}
\usepackage{booktabs}
\usepackage{newfloat}
\usepackage{listings}
\DeclareCaptionStyle{ruled}{labelfont=normalfont,labelsep=colon,strut=off} 
\floatstyle{ruled}
\newfloat{listing}{tb}{lst}{}
\floatname{listing}{Listing}
\title{Removing Rain Streaks via Task Transfer Learning}
\author{
    Written by AAAI Press Staff\textsuperscript{\rm 1}\thanks{With help from the AAAI Publications Committee.}\\
    AAAI Style Contributions by Pater Patel Schneider,
    Sunil Issar,\\
    J. Scott Penberthy,
    George Ferguson,
    Hans Guesgen,
    Francisco Cruz\equalcontrib,
    Marc Pujol-Gonzalez\equalcontrib
}
\affiliations{
    \textsuperscript{\rm 1}Association for the Advancement of Artificial Intelligence\\

    1900 Embarcadero Road, Suite 101\\
    Palo Alto, California 94303-3310 USA\\
    publications23@aaai.org
}

\usepackage{bibentry}

\begin{document}

\maketitle

\begin{abstract}
Due to the difficulty in collecting paired real-world training data, image deraining is currently dominated by supervised learning with synthesized data generated by e.g., Photoshop rendering. However, the generalization to real rainy scenes is usually limited due to the gap between synthetic and real-world data. In this paper, we first statistically explore why the supervised deraining models cannot generalize well to real rainy cases, and find the substantial difference of synthetic and real rainy data. Inspired by our studies, we propose to remove rain by learning favorable deraining representations from other connected tasks. In connected tasks, the label for real data can be easily obtained. Hence, our core idea is to learn representations from real data through task transfer to improve deraining generalization. We thus term our learning strategy as \textit{task transfer learning}. If there are more than one connected tasks, we propose to reduce model size by knowledge distillation. The pretrained models for the connected tasks are treated as teachers, all their knowledge is distilled to a student network, so that we reduce the model size, meanwhile preserve effective prior representations from all the connected tasks. At last, the student network is fine-tuned with minority of paired synthetic rainy data to guide the pretrained prior representations to remove rain. Extensive experiments demonstrate that proposed task transfer learning strategy is surprisingly successful and compares favorably with state-of-the-art supervised learning methods and apparently surpass other semi-supervised deraining methods on synthetic data. Particularly, it shows superior generalization over them to real-world scenes.
\end{abstract}

\section{Introduction}

\label{sec:intro}

\begin{figure}[t]
\centering
\begin{minipage}{0.32\linewidth}
	\centering{\includegraphics[width=1\linewidth]{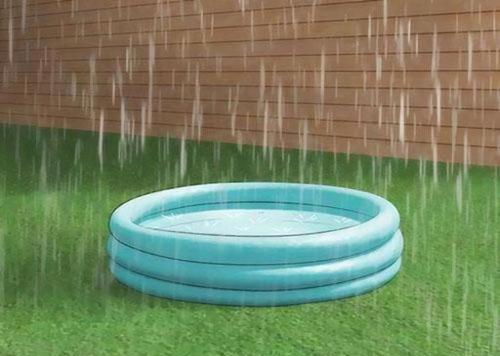}}
	\centerline{(a) Input}
\end{minipage}
\hfill
\begin{minipage}{0.32\linewidth}
	\centering{\includegraphics[width=1\linewidth]{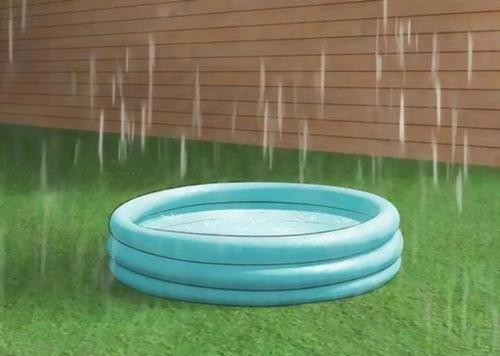}}
	\centerline{(b) DGUNet}
\end{minipage}
\hfill
\begin{minipage}{0.32\linewidth}
	\centering{\includegraphics[width=1\linewidth]{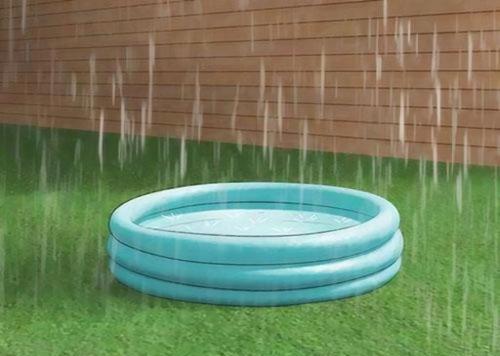}}
	\centerline{(c) MAXIM}
\end{minipage}
\vfill
\begin{minipage}{0.32\linewidth}
	\centering{\includegraphics[width=1\linewidth]{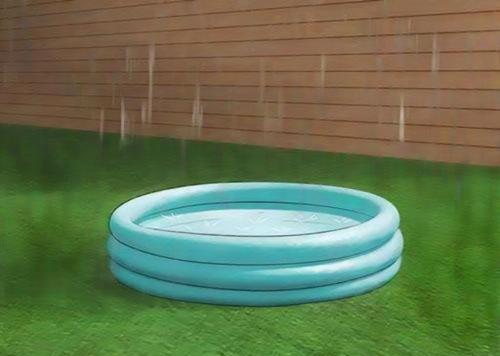}}
	\centerline{(d) MOEDN}
\end{minipage}
\hfill
\begin{minipage}{0.32\linewidth}
	\centering{\includegraphics[width=1\linewidth]{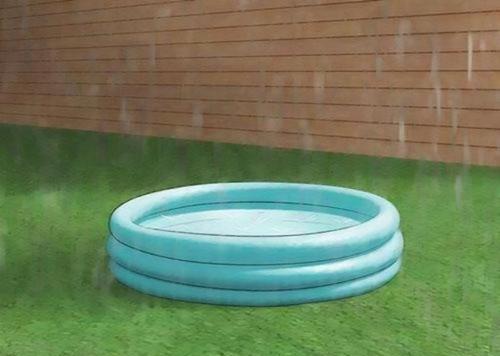}}
	\centerline{(e) VRGNet}
\end{minipage}
\hfill
\begin{minipage}{0.32\linewidth}
	\centering{\includegraphics[width=1\linewidth]{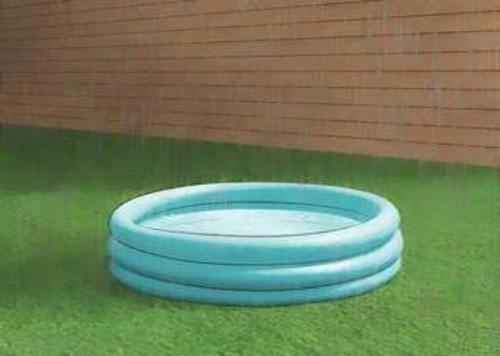}}
	\centerline{(f) Ours}
\end{minipage}
\caption{This figure illustrates that the representation from real data can better generalize to real-world scenes, and handle challenging rainy cases. (a) Input rainy images. (b)(c) Results of supervised DGUNet~\cite{mou-cvpr22-deep} and MAXIM \cite{tu-cvpr22-maxim}. (d)(e) Results of semi-supervised MOEDN \cite{huang-cvpr21-memory} and VRGNet \cite{wang-cvpr21-from}. (f) Our method.}
\label{fig:example}
\end{figure}


The performances of outdoor computer vision systems, e.g., autonomous driving, are inevitably impacted by rainy weather. Deraining, the removal of rain from images, has shown its potentials in improving the performance of other downstream computer vision tasks, especially for those in the applications of intelligent transportation and autonomous driving \cite{yang-tip20-advancing}. Conventional deraining methods typically optimize a model or learn an over-complete sparse dictionary to separate rain from rainy images \cite{wang-tip17-a,wang-icip16-a,li-cvpr16-rain}. Their performances are usually not satisfactory due to the limited capacity of the hand-crafted descriptors or models.

By exploring deep features, convolutional neural networks (CNNs) and vision transformers achieve remarkable progress in deraining task \cite{zhang-cvpr18-density,li-eccv18-recurrent,yang-pami19-joint,li-cvpr19-heavy,wang-cvpr19-spatial,wang-eccv20-rethinking,wang-tmm20-deep,valanarasu-cvpr22-transweather}. Currently, deraining networks are commonly trained under the supervision of reference ground truth images. Supervised training makes a network learn effective feature representation from input rainy images to fit their ground truth images under certain evaluation metrics.

Despite the success of supervised training methods, there exist two inevitable downsides. First, to achieve satisfactory performance, supervised training needs expensive simulation of rainy images based on ground truths. Second, the substantial gap between synthetic and real-world data intrinsically limits the generalization of supervised methods to real rainy scenes.

Recently, several works propose to improve generalization by using real rainy data in an semi-supervised manner \cite{huang-cvpr21-memory,liu-iccv21-unpaired,wang-cvpr21-from,yasarla-cvpr20-syn2real,ye-cvpr21-closing}. The basic idea of these methods is to generate pseudo label for real data under the guidance of synthetic data. Because the generated pseudo labels are usually inaccurate, these methods still cannot generalize well to real scenes.

Another way to improve generalization is self-supervised learning which usually learns representations from real data by designing proper surrogate task. However, it is usually hard to design effective surrogate task for deraining, in which the pseudo label must be derived from the real data itself. In this paper, we try to get rid of this constraint and propose task transfer learning to improve generalization also by using real rainy data. Specifically, for a given target task $\mathcal{T}_{tar}$, we attempt to find its connected tasks $\mathcal{T}_{i}, i=1, 2, ..., N$, and learn $\mathcal{T}_{tar}$-favorable prior representations from these connected tasks rather than from $\mathcal{T}_{tar}$. In other words, we transfer the solution of target task $\mathcal{T}_{tar}$ to solving its connected tasks, so that effective prior knowledge can be learned from real data. Hence, we term our learning strategy as \textit{task transfer learning}, and name the connected tasks as \textit{transfer tasks}. Under the significance of improving generalization, a transfer task should possess the following properties: 1) it uses real data as input to ensure better generalization of target task; 2) accurate labels for real data can be obtained easily to overcome the lacking of ground truth for real data in the target task; 3) it should be helpful/effective to the solution of target task. Therefore, self-supervised learning is just a special case of our task transfer learning, in which the pseudo labels are derived from real input images themselves.

Under the condition of having more than one transfer tasks, i.e., $N>1$, the model size increases proportionally with $N$ when we use pretained prior representations by direct concatenating method. We propose to treat these transfer tasks $\mathcal{T}_{i}, i=1, 2, ..., N$ as teachers, and all the prior knowledge in them is transferred to a student network $\mathcal{T}_{stu}$ via an effective knowledge distillation. By this way, we reduce model size as well as preserve prior knowledge from pretrained transfer tasks. At last, the knowledge in $\mathcal{T}_{stu}$ is fine-tuned by minority of synthetic pairs of $\mathcal{T}_{tar}$ to guide prior representations to solve target task $\mathcal{T}_{tar}$.

In this paper, deraining is our target task $\mathcal{T}_{tar}$. We first attempt to find the substantial reason for the limited generalization of supervised deraining models. Through analyzing the statistical difference between real-world and synthetic rainy data, an intrinsic cause leading to low generalization is discovered. It inspires future researches in this field, and also highlights the importance of using real data in image recovery tasks.

Allowing for the fact that the goal of our task transfer learning is to learn effective representations from real data to improve generalization, an accurate real rainy dataset is important. To the best of our knowledge, existing real rainy datasets are mainly collected from Internet, which is always not accurate to reflect real-world rainy scenes. Hence, we build a new real rainy dataset (named as RealRain) purely via capturing various rainy scenes with a camera to ensure that all data represents true distribution of real rainy scenes.

After obtaining accurate real-world dataset, another important aspect is to find transfer tasks $\mathcal{T}_{i}, i=1, 2, ..., N$ for our target deraining task $\mathcal{T}_{tar}$. Generally speaking, there are many methods to find effective transfer tasks for a given target task. The following two strategies can be usually used: 1) decomposing the target task into several sub-tasks, some of which can be selected as transfer tasks; 2) exploring the correlation/connection of target task to other tasks, e.g., deraining involves blurry image detail recovery, or the reconstruction of occluded objects. In later sections, we show how to find effective transfer tasks for deraining to learn generalization-favorable representations from real data. Figure \ref{fig:example} shows a deraining example, illustrating the better generalization of our task transfer learning to real rainy scenes.

Our contributions are summarized in the following:
\begin{itemize}
    \item We propose a new learning mechanism, i.e., task transfer learning, to learn highly effective generalization-favorable prior representations from real data and give out memory/time-efficient implementation strategy.
    \item We deeply explore the reason for the limited generalization of supervised deraining models via a quantitative statistical manner, which offers a new perspective for the future researches in this field.
    \item We build a new real rainy dataset purely via capturing various rainy scenes with a camera, which represents accurate distribution of rainy images, and facilitates further researches on the generalization of deraining models.
    \item We develop two simple yet effective transfer tasks for deraining to learn generalization-favorable prior representations from real data. Extensive experiments show the superiority of our task transfer deraining method over state-of-the-art supervised/semi-supervised methods on real rainy scenes.
\end{itemize}


\section{Related Work}
\label{sec:rela_works}
Our work aims at removing rain streaks from single images by using a transformer-based architecture, we summarize related works as follows.
\subsection{Deraining Works}
\noindent\textbf{Conventional optimization method.} Dictionary learning \cite{mairal-jmlr10-online} is originally utilized to solve the negative impact of rain streaks on the background, in which heuristic feature descriptors are used to recognize non-rain dictionary atoms \cite{wang-tip17-a,wang-icip16-a,kang-tip12-automatic,chen-tcsvt14-visual,huang-icme12-context,huang-tmm14-self}. 

\noindent\textbf{CNN-based method.} CNN-based deraining methods can be usually classified into direct mapping, residual based, and scattering model based methods. Direct mapping methods directly estimate rain-free images from their rainy versions. The representative works include SPANet \cite{wang-cvpr19-spatial}, TUM \cite{chen-cvpr22-learning}, DGUNet \cite{mou-cvpr22-deep}, AirNet \cite{li-cvpr22-all}. Residual based methods formulate a rainy image as the summation of the background layer and rain layers. A network is used to estimate easier rainy layer, the more complex background is obtained by subtracting estimated rainy layer from original rainy image.  This kind of methods includes MPRNet \cite{zamir-cvpr21-multi-stage}, MSPFN \cite{jiang-cvpr20-multi-scale}, DDN \cite{fu-cvpr17-removing}, JORDER \cite{yang-cvpr17-deep,yang-pami19-joint}, DID-MDN \cite{zhang-cvpr18-density}, RESCAN \cite{li-eccv18-recurrent}, DAF-Net \cite{hu-cvpr19-depth}, PReNet \cite{ren-cvpr19-progressive}, RCDNet \cite{wang-cvpr20-a}. Scattering model based methods render and learn atmospheric light and transmission of vapor to remove rain streaks as well as vapor effect, e.g., PYM+GAN \cite{li-cvpr19-heavy}, ASV-joint \cite{wang-eccv20-rethinking}.

\noindent\textbf{Semi-supervised Methods.} Semi-supervised methods are proposed to improve generalization of deraining models to real rainy scenes by using real data. The basic idea of these methods is to generate pseudo label for real data under the guidance of synthetic data, e.g., MOEDN \cite{huang-cvpr21-memory}, UDRDR \cite{liu-iccv21-unpaired}, VRGNet \cite{wang-cvpr21-from}, Syn2Real \cite{yasarla-cvpr20-syn2real}, JRGR \cite{ye-cvpr21-closing}. Because the generated pseudo labels are usually inaccurate, these methods still cannot generalize well to real scenes.

\noindent\textbf{Transformer/MLP-based Methods.} Recently, some novel architectures, e.g. transformer and MLP, are proposed to solve low-level vision tasks by using their long-rang modelling capacities to overcome the inductive biases of CNN-based architecture. For deraining, these works includes KiT \cite{lee-cvpr22-knn}, MAXIM \cite{tu-cvpr22-maxim},  TransWeather \cite{valanarasu-cvpr22-transweather}, Uformer \cite{wang-cvpr22-uformer}, Restormer \cite{zamir-cvpr22-restormer}. Our work is also based on transformer, but we adaptively fuse global dependency and local details of images to obtain better recovery quality. 

\section{Method}
\label{sec:self_algo}

Existing deraining networks are commonly trained in supervised manner based on synthetic training pairs. These synthesized rainy images usually take lots of efforts and most importantly, cannot represent the true attributes of real-world rainy images, leading to limited generalization capacity. In this paper, we deeply explore the difference between synthetic and real-world data and discover the substantial reason for the limited generalization of supervised deraining models. In order to train a model generalizing well to real scenes via our proposed task transfer learning, we analyze the deraining mechanism of supervised training strategy and find two highly effective transfer tasks to learn prior deraining representations from real rainy/non-rainy data. To reduce the model size during the deraining process, we regard transfer tasks as teachers, and distill the knowledge in them to a student network. Then, the taught student network is guided to remove rain with minority of synthetic rainy pairs.

\begin{figure}[t]
\centering
\begin{minipage}{1.0\linewidth}
\centering{\includegraphics[width=1\linewidth]{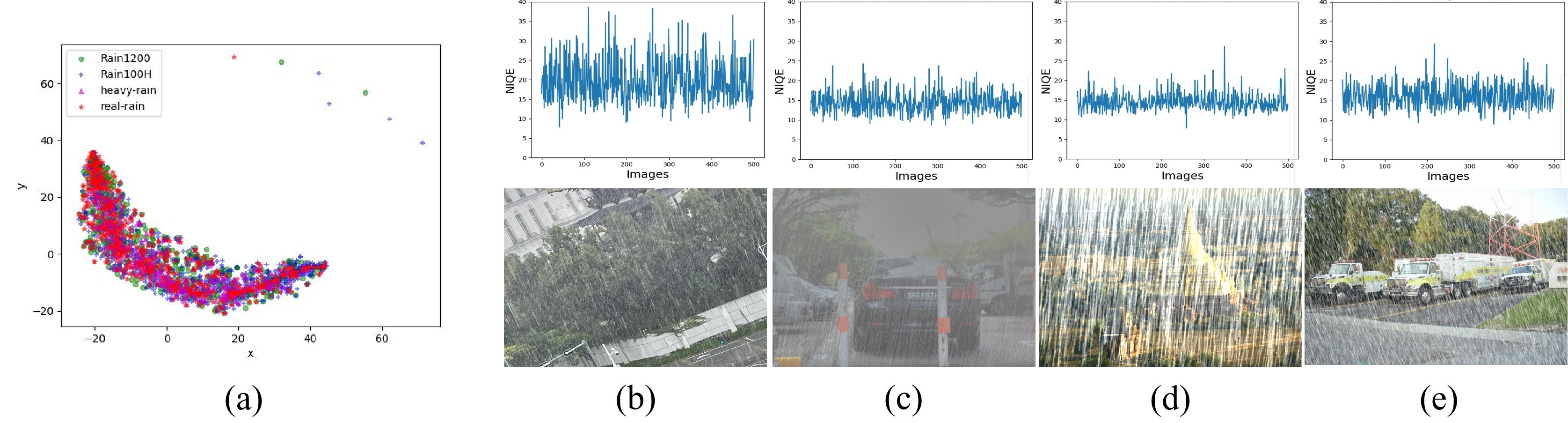}}
\end{minipage}
\caption{The differences between real and synthetic rainy images. (a) Comparing t-SNE distributions in terms of real and synthetic rainy data. (b) NIQE values of $500$ real-world rainy images. (c) (d) (e) NIQE values of $500$ synthetic rainy images from ``heavy-rain'', Rain100H, and Rain1200, respectively. We observe that on average, the NIQE value of real rainy images is significantly higher than that of synthetic rainy images, even the synthetic rainy images having very heavy rain streaks, e.g. (d).}
\label{fig:real-fake-compare}
\end{figure}

\subsection{Limited Generalization of Supervised Models}
Currently, the low generalization of supervised deraining models is always attributed to the limited amount of training pairs covering only part of real rainy patterns or the distribution gap between synthetic and real-world rainy images. In this section, we give a more specific study on the substantial difference of real and synthetic rainy datasets. Figure \ref{fig:real-fake-compare} (a) presents the t-SNE \cite{maaten-jmlr08-visualizing} distribution relationships with respect to real and synthetic rainy images, in which three representative synthetic datasets are selected, i.e., ``heavy-rain'' \cite{li-cvpr19-heavy}, Rain100H \cite{yang-cvpr17-deep} and Rain1200 \cite{zhang-cvpr18-density}. The real rainy images are randomly selected from our newly-built RealRain. We observe that the distribution of synthetic rainy images is very similar to that of real rainy images. Moreover, the distribution of synthetic rainy images covers that of real rainy images, which illustrates that the synthetic rainy patterns nearly contains that of real rain. Therefore, our experiment proves that the distribution gap between real and synthetic rainy images and the lacking of complete rainy patterns in synthetic dataset are not the substantial reasons for the limited generalization of supervised deraining models.

\begin{figure*}[t]
\centering
\begin{minipage}{1\linewidth}
\centering{\includegraphics[width=1\linewidth]{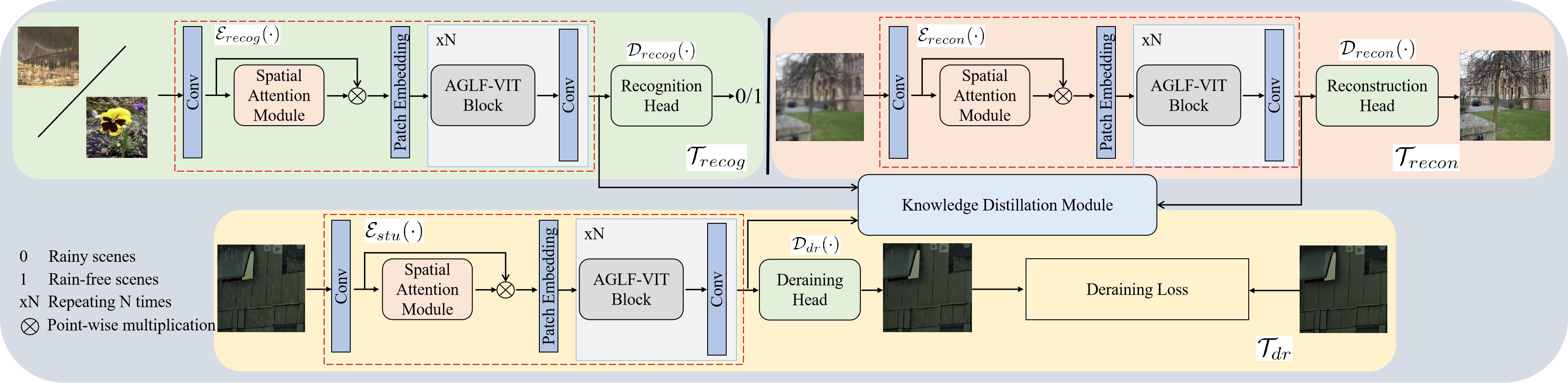}}
\end{minipage}
\caption{Overall architecture of our method. $\mathcal{E}_{recog}(\cdot)$ and $\mathcal{D}_{recog}(\cdot)$ are for the recognition task $\mathcal{T}_{recog}$, and $\mathcal{E}_{recon}(\cdot)$ and $\mathcal{D}_{recon}(\cdot)$ are for the reconstruction task $\mathcal{T}_{recon}$. These two tasks are trained independently. After their training, the knowledge in $\mathcal{E}_{recog}(\cdot)$ and $\mathcal{E}_{recon}(\cdot)$ is distilled to a student network $\mathcal{E}_{stu}(\cdot)$, and $\mathcal{D}_{dr}(\cdot)$ is followed to remove rain streaks.}
\label{fig:framework}
\end{figure*}

Natural Image Quality Evaluation (NIQE) \cite{mittal-spl12-making} measures the deviation of degraded images from natural high-quality images, i.e., the damage degree of degradation factor (rain in this paper) to natural images. We show NIQE of real-world and synthetic rainy images in Figure \ref{fig:real-fake-compare}. It is observed that the NIQE values of synthetic rainy images are consistently much lower than these of real ones, which illustrates that the damage of real rain to image quality is much significant than the synthetic rain streaks, even when the synthetic rain looks heavier than real rain, e.g., very heavy rain in ``heavy-rain'' dataset, and the very bright rain streaks in Rain100H. Hence, we conclude that the different damaging degree of real and synthetic rain to high-quality image may be the substantial reason for the limited generalization of supervised deraining models. The damage degree is irrelevant to the amount of rainy patterns and the data distribution. It is related to the synthesis method of rainy images. According to the imaging mechanism of a camera, rain and background are first captured by sensors and form raw data. The final RGB image is obtained by a series of non-linear transformations to the raw data. While, synthetic rainy images are synthesized only by some simple combination of background and rain layers, e.g., linear combination.

We would like to point out that our study about the reason for the limited generalization of supervised deraining model is an important progress in deraining field. 1) It corrects some false cognition about the reason for the limited generalization, e.g., insufficient training pairs, and lacking complete real rainy patterns. These misunderstandings may mislead some researcher to pay their attentions to some trivial aspects, e.g., spending more time in pursuing more rainy patterns to build very large-scale dataset. 2) It guides some researchers to find better synthesis method for synthetic rainy images to build generalization-favoured training dataset, i.e., making rendered rain layer generate similar damage degree to high-quality images as real rain. 3) Under the condition of lacking effective training data, our study further proves the importance of using real rainy data to improve generalization, because synthetic rain is different from real rain in changing high-quality images.

\subsection{Finding Effective Transfer Tasks for Deraining}
\label{sec:trans_task}
We have studied the substantial reason for the limited generalization of supervised deraining models. In this paper, we focus on finding effective transfer tasks for deraining to improve generalization by using real data.


By observing real rainy images, we find that the image degradation caused by rain mainly include two types: 1) rain streaks occlude the content behind them, and 2) rain streaks blur image details due to their scattering to light. Thus, we factorize the whole deraining task $\mathcal{T}_{dr}$ into the combination of two subtasks (transfer tasks): 1) learning discriminative representation of rainy and rain-free contents by recognizing rainy or non-rainy scenes, denoted as $\mathcal{T}_{recog}$, which facilitates the deraining decoder to restrain rainy features and preserve/extend rain-free ones to remove rain streaks, so that the first type of occlusion degradation can be solved and recover the contents behind rain streaks, and 2) learning detail-recovering representation from blurry images by reconstructing blur degradation images, denoted as $\mathcal{T}_{recon}$, which helps decoder recover the lost image details to solve the second type of blur degradation. To reduce the model size, the learned prior knowledge from $\mathcal{T}_{recog}$ and $\mathcal{T}_{recon}$ is distilled to a student network. Then, the student network is fine-tuned by minority of synthetic rainy images to guide the pretrained representation to remove rain. The whole architecture is shown in Figure \ref{fig:framework}.


\subsection{Discriminative Representation Learning}
\label{sec:recog}
Transfer task $\mathcal{T}_{recog}$ is targeted to learn discriminative representation for recognizing rain and rain-free contents, which is realized via a transformer based encoder-decoder architecture. We denote the encoder and decoder for $\mathcal{T}_{recog}$ as $\mathcal{E}_{recog}(\cdot)$ and $\mathcal{D}_{recog}(\cdot)$, respectively. Their architectures are in Figure \ref{fig:framework}. $\mathcal{T}_{recog}$ uses real data as follows: the input for $\mathcal{E}_{recog}(\cdot)$ is real rainy or rain-free images and $\mathcal{D}_{recog}(\cdot)$ outputs 0/1 standing for rainy/rain-free scenes. Note that the training for $\mathcal{T}_{recog}$ is a supervised process. However, the labels for rainy and rain-free scenes are easy to obtain and the recognition labels are accurate, which ensures the learning of better representation than directly using paired inaccurate synthetic rainy images. The loss function for $\mathcal{T}_{recog}$ is:
\begin{equation}\label{eq:recog-loss}
 \mathcal{L}_{recog}=\sum_{i=1}^{M}\Vert o^{i}_{recog}-l^{i}_{recog} \Vert^2,
\end{equation}
where $o^{i}_{recog}$ is the output of $\mathcal{D}_{recog}(\cdot)$ for the $i$-th input sample and the ground truth $l^i_{recog} \in \{0, 1\}$ denotes rainy or rain-free, $M$ is the number of training samples.

In above section, $\mathcal{T}_{recog}$ is selected as a transfer task for $\mathcal{T}_{dr}$ by analyzing the degradation of rain to images. Here, we further support the rationality of $\mathcal{T}_{recog}$ as a transfer task by showing a fact that deraining process actually implies recognizing rain and rain-free scenes. This fact is easily verified: when inputting a rainy image to a good deraining model, a rain-removed result will be output. However, when the input is a clear image, it will not be changed by the model and output intactly. This fact indicates that the network also learns to discriminate rainy and rain-free images during supervised deraining process, which motivates our recognition scheme.

\subsection{Detail-Recovering Representation Learning}
\label{sec:recons}

Transfer task $\mathcal{T}_{recon}$ is to learn detail-recovering representation to solve the blur caused by the scattering of rain streaks to light, in which a transformer based encoder-decoder architecture is used, as shown in Figure \ref{fig:framework}. The encoder and decoder are denoted as $\mathcal{E}_{recon}(\cdot)$ and $\mathcal{D}_{recon}(\cdot)$, respectively. $\mathcal{T}_{recon}$ uses real data as follows: the input for $\mathcal{E}_{recon}(\cdot)$ is a blurry image, and $\mathcal{D}_{recon}(\cdot)$ predicts a corresponding clear image. The training process is also a supervised one, the blurry input can be easily obtained by blurring a clear image with a Gaussion kernel, the ground truth is the clear image itself. Similar to $\mathcal{T}_{recog}$, $\mathcal{T}_{recon}$ also learns generalization-favorable representation by using paired real training data. The loss is defined as:
\begin{equation}\label{eq:recon-loss}
\begin{aligned}
 \mathcal{L}_{recon}&= \sum_{i=1}^{M}(\Vert \mathbf{O}^{i}_{recon}-\mathbf{C}^{i}_{recon} \Vert^2 \\
 &+ \Vert \bigtriangledown\mathbf{O}^{i}_{recon}-\bigtriangledown\mathbf{C}^{i}_{recon} \Vert^2),
\end{aligned}
\end{equation}
where $\mathbf{O}^{i}_{recon}$ is the output for the $i$-th sample, $\mathbf{C}^{i}_{recon}$ is its corresponding ground truth, $\bigtriangledown$ represents the horizontal and vertical gradients, and $M$ is the number of training samples.

\subsection{Knowledge Distillation for Reducing Model Size}
\label{sec:kd}
The model size is large if we use the knowledge in $\mathcal{E}_{recog}(\cdot)$ and $\mathcal{E}_{recon}(\cdot)$ by directly concatenating their outputs. To address this issue, we regard $\mathcal{E}_{recog}(\cdot)$ and $\mathcal{E}_{recon}(\cdot)$ as teachers, and transfer their pretrained knowledge to a student network $\mathcal{E}_{stu}(\cdot)$ via knowledge distillation \cite{hinton-arxiv15-distilling}. Knowledge distillation is conducted on our real rainy dataset RealRain $\{\mathbf{I}_{i}\}_{i=1}^{M}$, which mainly includes two aspects. On one hand, we adopt direct representation matching, i.e., using the outputs of $\mathcal{E}_{recog}(\cdot)$ and $\mathcal{E}_{recon}(\cdot)$ to simultaneously supervise the output of $\mathcal{E}_{stu}(\cdot)$:
\begin{equation}\label{eq:kd1-loss}
\begin{aligned}
 \mathcal{L}_{kdd}&=\sum_{i=1}^{M}\Vert \mathcal{E}_{stu}(\mathbf{I}_{i}) - 
\mathcal{E}_{recog}(\mathbf{I}_{i}) \Vert^2 \\
&+ \Vert \mathcal{E}_{stu}(\mathbf{I}_{i}) - 
\mathcal{E}_{recon}(\mathbf{I}_{i}) \Vert^2,
\end{aligned}
\end{equation}
Direct representation matching may lead to feature smoothing due to the average convergence property of MSE loss. Hence, on the other hand, we borrow $\mathcal{D}_{recog}(\cdot)$ and $\mathcal{D}_{recon}(\cdot)$ to realize indirect representation matching:
\begin{equation}\label{eq:kd2-loss}
\begin{aligned}
 \mathcal{L}_{kdi}&=\sum_{i=1}^{M}\Vert \mathcal{D}_{recog}(\mathcal{E}_{stu}(\mathbf{I}_{i})) - 
\mathcal{D}_{recog}(\mathcal{E}_{recog}(\mathbf{I}_{i})) \Vert^2 \\
&+ \Vert \mathcal{D}_{recon}(\mathcal{E}_{stu}(\mathbf{I}_{i})) - 
\mathcal{D}_{recon}(\mathcal{E}_{recon}(\mathbf{I}_{i})) \Vert^2.
\end{aligned}
\end{equation}
The overall loss for knowledge distillation is as follows:
\begin{equation}\label{eq:kd-loss}
 \mathcal{L}_{kd} = \mathcal{L}_{kdd} + \mathcal{L}_{kdi}.
\end{equation}

\subsection{Guiding Student Network to Remove Rain}
After obtaining mature student network $\mathcal{E}_{stu}(\cdot)$ by knowledge distillation, we match it with a deraining decoder $\mathcal{D}_{dr}(\cdot)$ and guide it to remove rain via minority of synthetic rainy image pairs. During such process, $\mathcal{E}_{stu}(\cdot)$ is fine-tuned with a relatively small learning rate and $\mathcal{D}_{dr}(\cdot)$ is trained with a larger learning rate. The loss function is similar to that of detail-recovering representation learning, i.e., Eq. \eqref{eq:recon-loss}.

\begin{figure}[t]
\centering
\begin{minipage}{1\linewidth}
\centering{\includegraphics[width=0.8\linewidth]{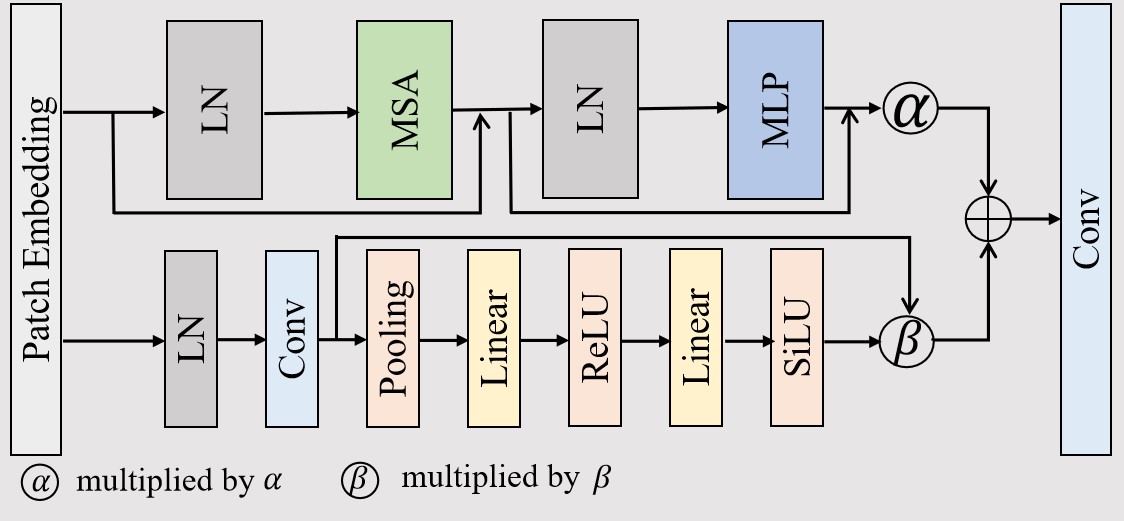}}
\end{minipage}
\caption{The architecture of our AGLF-ViT. $\alpha$ and $\beta$ are learnable parameters to adaptively combine global context dependency and local image detail features.}
\label{fig:aglf-vit}
\end{figure}

\subsection{Network Structure}
\label{sec:network}

As shown in Figure \ref{fig:framework}, $\mathcal{E}_{recog}(\cdot)$, $\mathcal{E}_{recon}(\cdot)$ and $\mathcal{E}_{stu}(\cdot)$ possess the same architecture, in which a convolutional layer is first used to project image into shallow feature space, then we use a spatial attention module to highlight rainy contents to facilitate their subsequent recognition and removal. $9$ transformer blocks are followed to extract deep features. Common self-attention mechanism usually models global context dependency. While, local image detail recovery is also important for image deraining. Inspired by \cite{liu-eccv22-ghost}, we design an \textbf{A}daptive \textbf{G}lobal and \textbf{L}ocal feature \textbf{F}usion \textbf{Vi}sion \textbf{T}ransformer block (AGLF-ViT) to adaptively combine the global context relationship modelled by self-attention and local image detail features extracted by CNN block, shown in Figure \ref{fig:aglf-vit}. The structures for $\mathcal{D}_{recon}(\cdot)$ and $\mathcal{D}_{dr}(\cdot)$ are the same, which contain an SPP module to diversify features in scale and a convolutional layer to output the results. The structure of $\mathcal{D}_{recog}(\cdot)$ includes SPP module, followed by an average pooling and a fully connected layer to output recognition result. More details are in supplement.

\begin{table}
\begin{center}
\begin{tabular}{c|c|c|c}
\multirow{2}{*}{Methods} & Test-I & Test-II & Test-III \\
                         & PSNR/SSIM & PSNR/SSIM & PSNR/SSIM \\
\toprule
DGUNet & 34.75/0.935 & 32.47/0.919 & \underline{35.83}/0.948 \\
MAXIM & \underline{34.87}/\underline{0.939} & \underline{33.01}/\underline{0.924} & 33.53/\underline{0.958} \\
Restormer & 34.68/0.933 & 32.91/0.923 & 35.34/0.946 \\
TUM & 32.04/0.902 & 30.72/0.892 & 22.75/0.855 \\
JRGR & 23.43/0.835 & 23.02/0.775 & 23.91/0.883 \\
MOEDN & 30.29/0.898 & 29.73/0.886 & 32.63/0.945 \\
VRGNet & 32.90/0.911 & 31.11/0.904 & 32.23/0.954 \\
\midrule
Ours & \textbf{35.01}/\textbf{0.941} & \textbf{33.23}/\textbf{0.938} & \textbf{36.65}/\textbf{0.960} \\
\bottomrule
\end{tabular}
\end{center}
\caption{PSNR and SSIM comparison with state-of-the-art methods on our three testing datasets. The best and second best numbers are in bold and underlined, respectively.}
\label{tab:psnr_ssim_sota}
\end{table}

\section{Experiments}
\label{sec:expe}

\begin{figure*}[!t]
\begin{center}
\begin{minipage}{0.095\linewidth}
\centering{\includegraphics[width=1\linewidth]{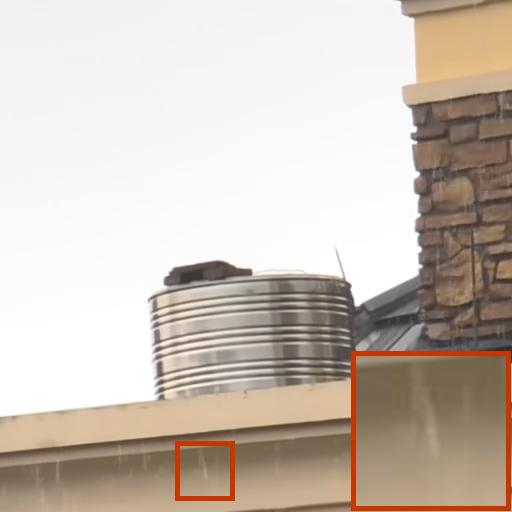}}
\centerline{(a)}
\end{minipage}
\hfill
\begin{minipage}{0.095\linewidth}
\centering{\includegraphics[width=1\linewidth]{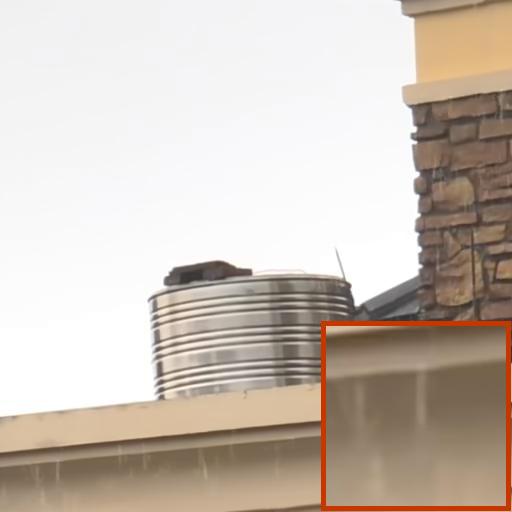}}
\centerline{(b)}
\end{minipage}
\hfill
\begin{minipage}{0.095\linewidth}
\centering{\includegraphics[width=1\linewidth]{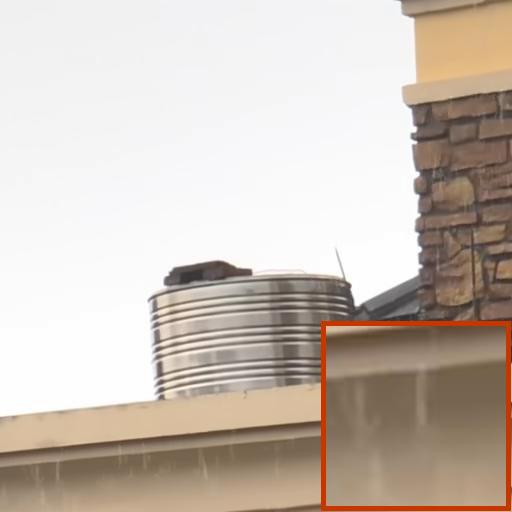}}
\centerline{(c)}
\end{minipage}
\hfill
\begin{minipage}{0.095\linewidth}
\centering{\includegraphics[width=1\linewidth]{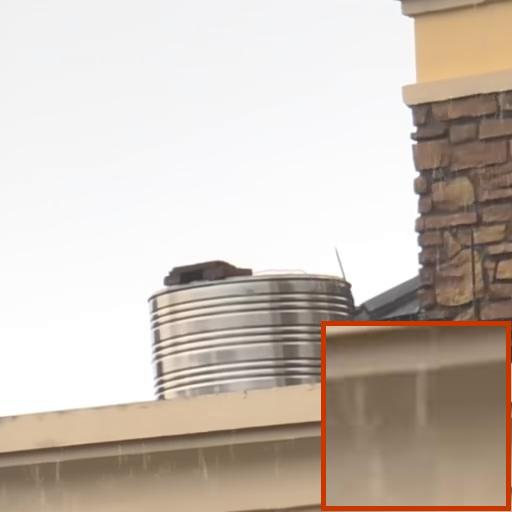}}
\centerline{(d)}
\end{minipage}
\hfill
\begin{minipage}{0.095\linewidth}
\centering{\includegraphics[width=1\linewidth]{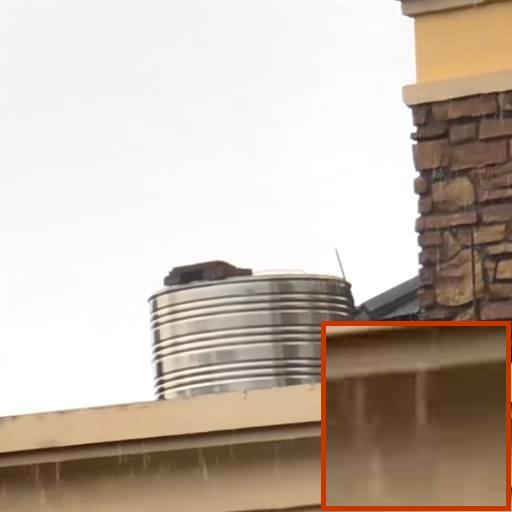}}
\centerline{(e)}
\end{minipage}
\hfill
\begin{minipage}{0.095\linewidth}
\centering{\includegraphics[width=1\linewidth]{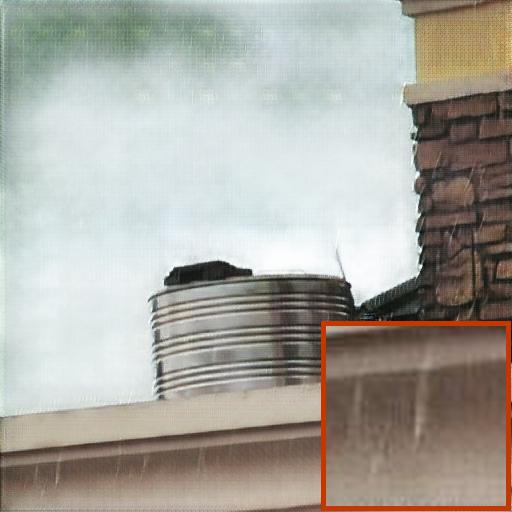}}
\centerline{(f)}
\end{minipage}
\hfill
\begin{minipage}{0.095\linewidth}
\centering{\includegraphics[width=1\linewidth]{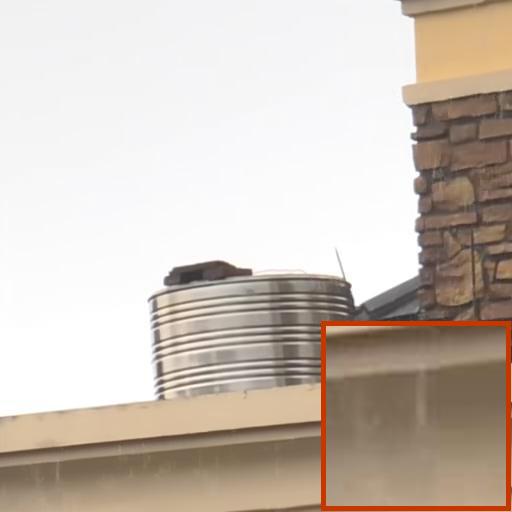}}
\centerline{(g)}
\end{minipage}
\hfill
\begin{minipage}{0.095\linewidth}
\centering{\includegraphics[width=1\linewidth]{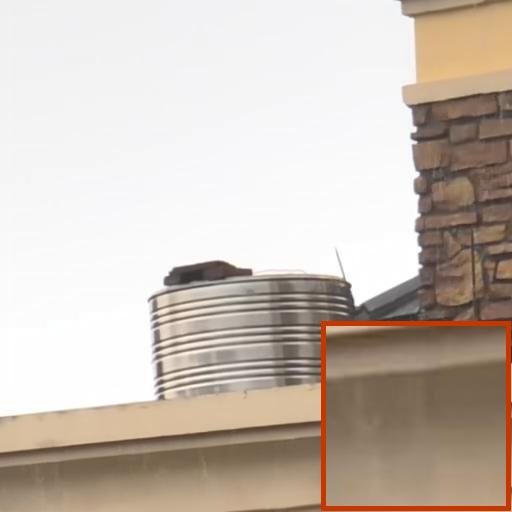}}
\centerline{(h)}
\end{minipage}
\hfill
\begin{minipage}{0.095\linewidth}
\centering{\includegraphics[width=1\linewidth]{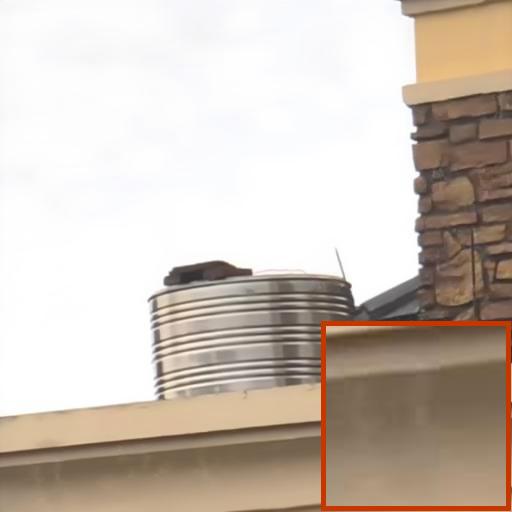}}
\centerline{(i)}
\end{minipage}
\hfill
\begin{minipage}{0.095\linewidth}
\centering{\includegraphics[width=1\linewidth]{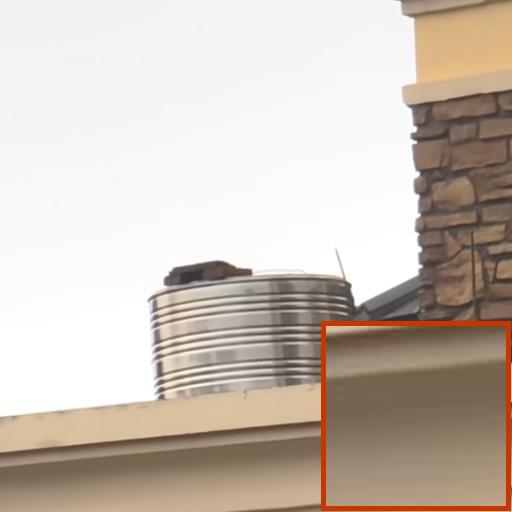}}
\centerline{(j)}
\end{minipage}
\end{center}
\caption{Qualitative comparisons on synthetic rainy image. (a) Input. (b)-(i) Results of DGUNet, MAXIM, Restormer, TUM, JRGR, MOEDN, VRGNet, and our method, respectively. (j) Ground Truth. The patch of the closeup is given in the first image.}
\label{fig:synthetic_compare}
\end{figure*}

\begin{figure*}[!t]
\begin{center}
\begin{minipage}{0.105\linewidth}
\centering{\includegraphics[width=1\linewidth]{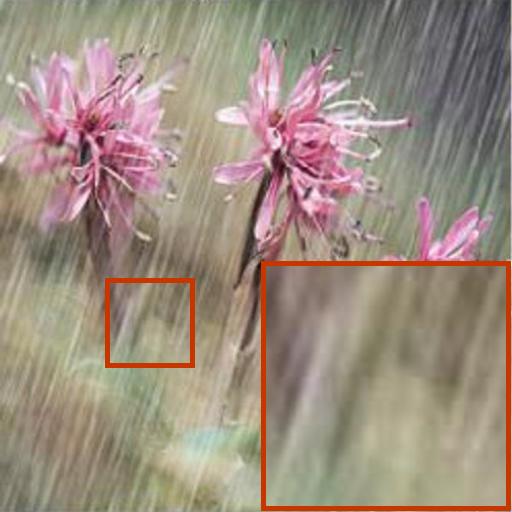}}
\end{minipage}
\hfill
\begin{minipage}{0.105\linewidth}
\centering{\includegraphics[width=1\linewidth]{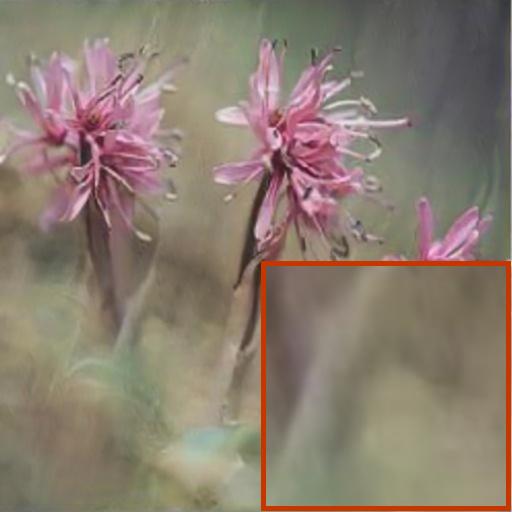}}
\end{minipage}
\hfill
\begin{minipage}{0.105\linewidth}
\centering{\includegraphics[width=1\linewidth]{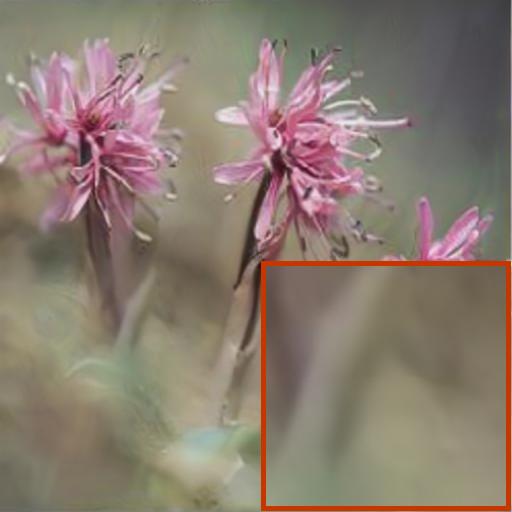}}
\end{minipage}
\hfill
\begin{minipage}{0.105\linewidth}
\centering{\includegraphics[width=1\linewidth]{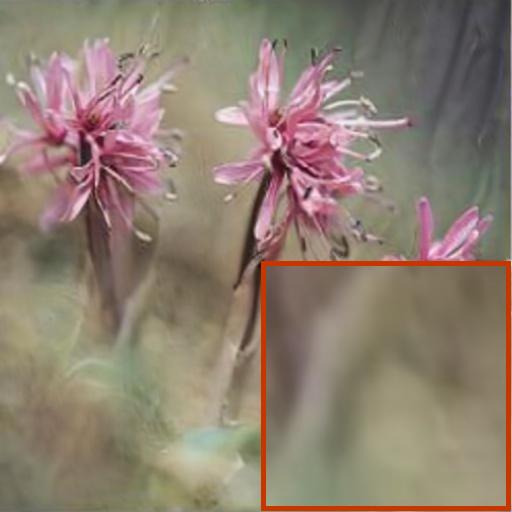}}
\end{minipage}
\hfill
\begin{minipage}{0.105\linewidth}
\centering{\includegraphics[width=1\linewidth]{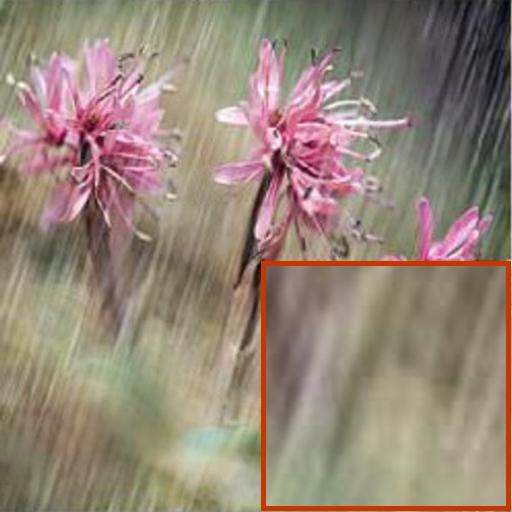}}
\end{minipage}
\hfill
\begin{minipage}{0.105\linewidth}
\centering{\includegraphics[width=1\linewidth]{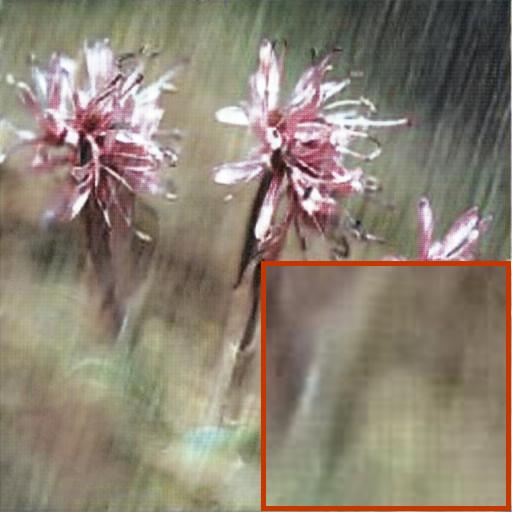}}
\end{minipage}
\hfill
\begin{minipage}{0.105\linewidth}
\centering{\includegraphics[width=1\linewidth]{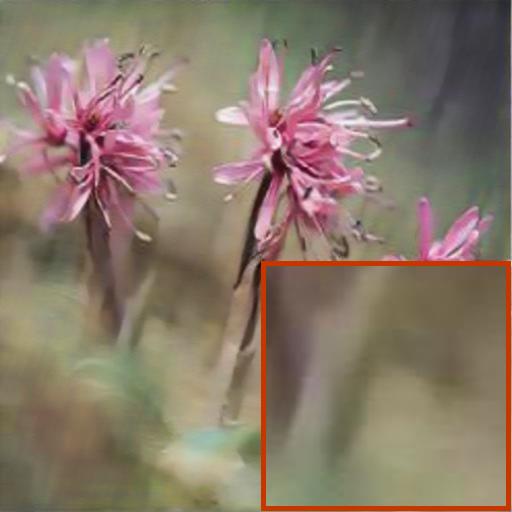}}
\end{minipage}
\hfill
\begin{minipage}{0.105\linewidth}
\centering{\includegraphics[width=1\linewidth]{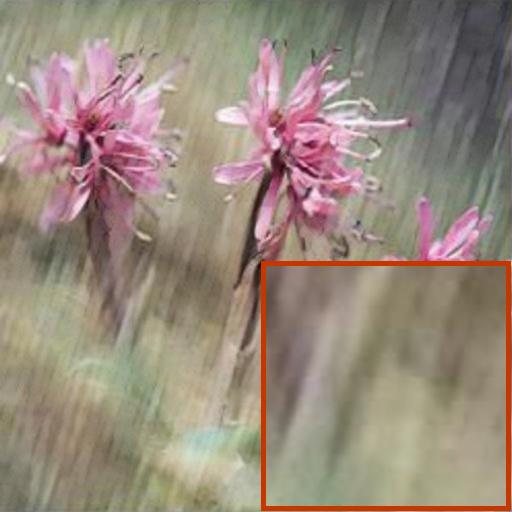}}
\end{minipage}
\hfill
\begin{minipage}{0.105\linewidth}
\centering{\includegraphics[width=1\linewidth]{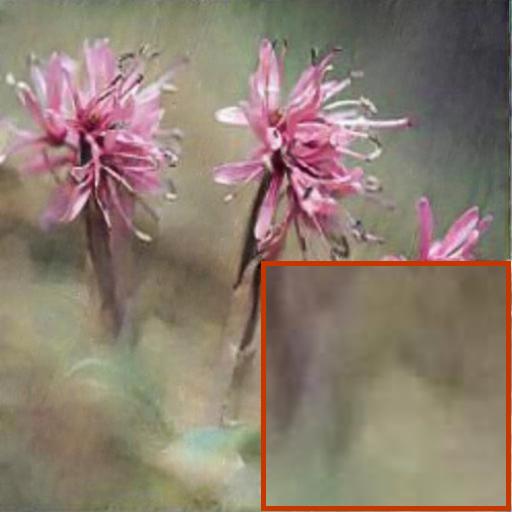}}
\end{minipage}
\vfill
\begin{minipage}{0.105\linewidth}
\centering{\includegraphics[width=1\linewidth]{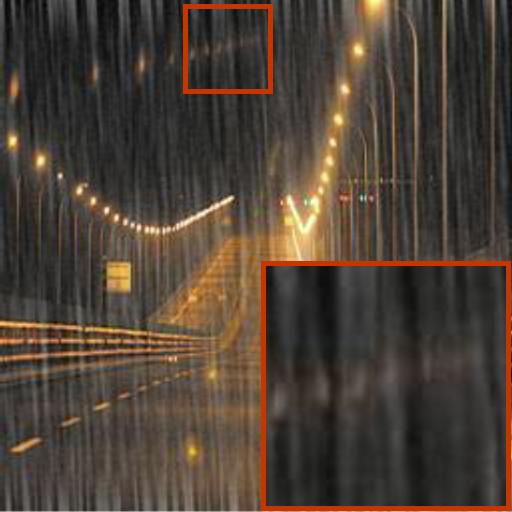}}
\centerline{(a)}
\end{minipage}
\hfill
\begin{minipage}{0.105\linewidth}
\centering{\includegraphics[width=1\linewidth]{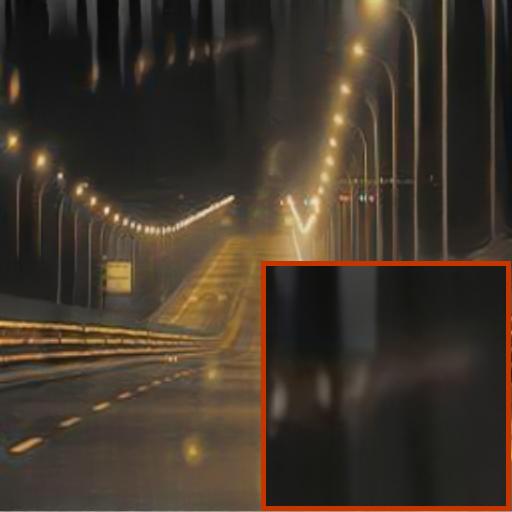}}
\centerline{(b)}
\end{minipage}
\hfill
\begin{minipage}{0.105\linewidth}
\centering{\includegraphics[width=1\linewidth]{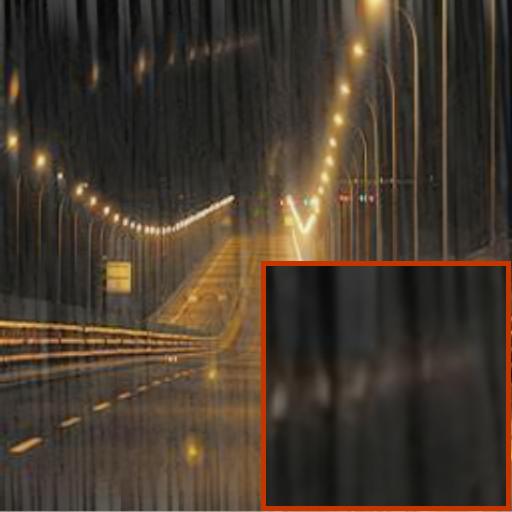}}
\centerline{(c)}
\end{minipage}
\hfill
\begin{minipage}{0.105\linewidth}
\centering{\includegraphics[width=1\linewidth]{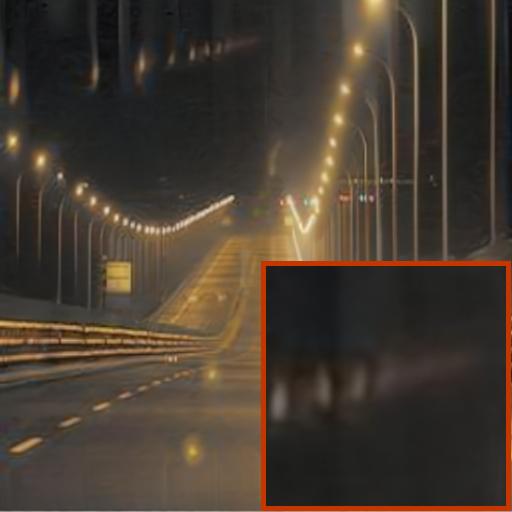}}
\centerline{(d)}
\end{minipage}
\hfill
\begin{minipage}{0.105\linewidth}
\centering{\includegraphics[width=1\linewidth]{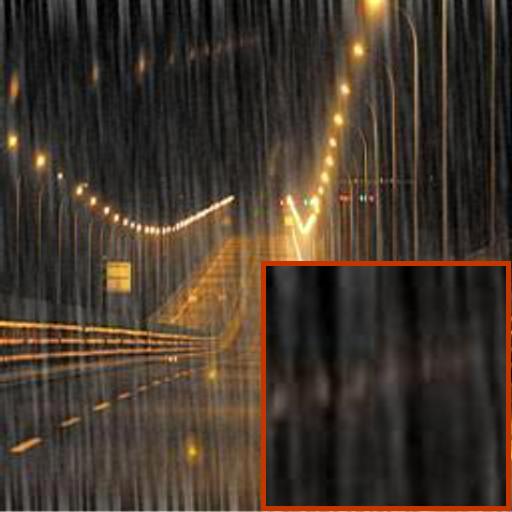}}
\centerline{(e)}
\end{minipage}
\hfill
\begin{minipage}{0.105\linewidth}
\centering{\includegraphics[width=1\linewidth]{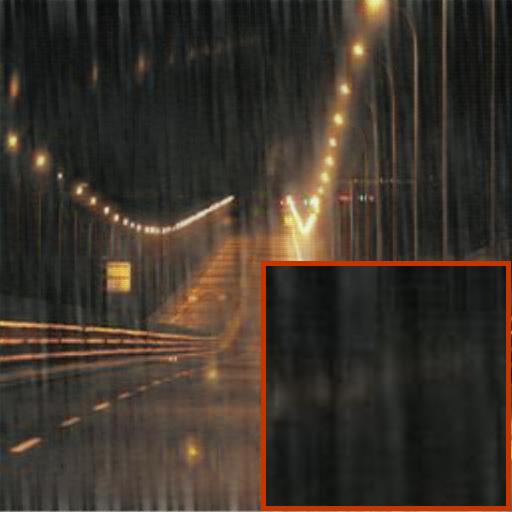}}
\centerline{(f)}
\end{minipage}
\hfill
\begin{minipage}{0.105\linewidth}
\centering{\includegraphics[width=1\linewidth]{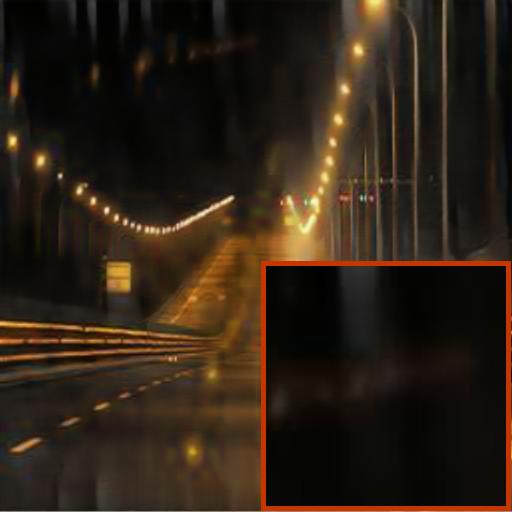}}
\centerline{(g)}
\end{minipage}
\hfill
\begin{minipage}{0.105\linewidth}
\centering{\includegraphics[width=1\linewidth]{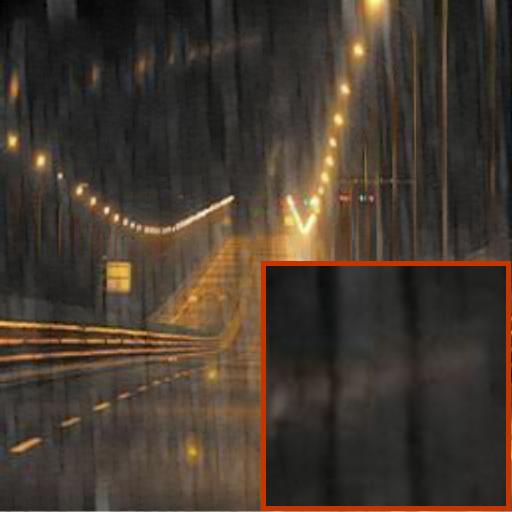}}
\centerline{(h)}
\end{minipage}
\hfill
\begin{minipage}{0.105\linewidth}
\centering{\includegraphics[width=1\linewidth]{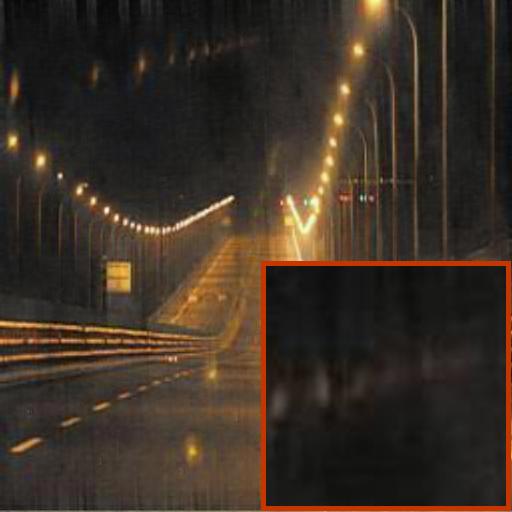}}
\centerline{(i)}
\end{minipage}
\end{center}
\caption{Qualitative comparison on real-world rainy images. (a) Input. (b)-(i) Results of DGUNet, MAXIM, Restormer, TUM, JRGR, MOEDN, VRGNet, and our method, respectively.}
\label{fig:practical_compare}
\end{figure*}

\begin{table*}[t]
\begin{center}
\begin{tabular}{c|c|c|c|c|c|c|c|c}
\toprule
Methods & DGUNet & MAXIM & Restormer & TUM & JRGR & MOEDN & VRGNet & Ours \\
\midrule
    Total score  & $8901$ & $9074$ & $8324$ & $6475$ & $6000$ & $11151$ & $7426$ & $14649$ \\ 
\bottomrule
\end{tabular}
\end{center}
\caption{Total score comparisons of different methods in user-study.}
\label{tab:user-study}
\end{table*}

In this section, we evaluate our method and compare it with recent works on synthetic and real data. 

\noindent\textbf{Training details.} During training, all the patches are randomly cropped from the original images with a fixed size of $256\times256$. Adam \cite{kingma-iclr15-adam} is selected as our optimizer. The initial learning rate for $\mathcal{T}_{recog}$ and $\mathcal{T}_{recon}$ are $0.0004$. The initial fine-tuning learning rate for $\mathcal{E}_{stu}(\cdot)$ is $0.00004$, and that for $\mathcal{D}_{dr}$ is $0.0004$. All the learning rates are decayed by multiplying $0.5$ when the losses do not decrease. Our code is implemented on a NVIDIA v100NV32 GPU based on Pytorch. More details are in supplement.

\noindent\textbf{Datasets.} To facilitate the researches on the generalization of deraining models, we collect a real-world rainy dataset (RealRain) purely by capturing diverse rainy scenes by a camera. Our RealRain contains $1500$ samples so far, covering light, medium and heavy rain conditions. Its rainy scenes include buildings, plants, streets, pedestrian, etc. Our real non-rain dataset (RealClear) contains $10000$ samples selected from the ground truth of Rain1200 \cite{zhang-cvpr18-density}, and Rain2800 \cite{fu-cvpr17-removing}. Moreover, we randomly select $6000$ training pairs from the training dataset of \cite{zhang-cvpr18-density} to guide the student network $\mathcal{E}_{stu}(\cdot)$ to remove rain. Because our work is to improve the generalization of deraining model to real scenes, we constitute our testing datasets as follows: 1) we follow \cite{wang-eccv20-rethinking} and randomly select $100$ synthetic pairs from Rain1200, $100$ from Rain2800, and $100$ from Rain800 \cite{zhang-arxiv17-image} to construct our first testing dataset to cover more synthetic rainy types, named Test-I (300 pairs); 2) the ``rain streaks'' testing dataset in \cite{li-cvpr19-single} well renders the influence of rain streaks on background, and is solely selected as our second testing dataset, named Test-II; 3) the paired real rain dataset generated from rainy video sequences \cite{wang-cvpr19-spatial} is closer to real-world rainy scenes, which can better verify the generalization of rainy models and is selected as our third testing testing dataset, named Test-III. Recent SOTA supervised models DGUNet \cite{mou-cvpr22-deep}, MAXIM \cite{tu-cvpr22-maxim}, Restormer \cite{zamir-cvpr22-restormer}, TUM \cite{chen-cvpr22-learning} and semi-supervised methods JRGR \cite{ye-cvpr21-closing}, MOEDN \cite{huang-cvpr21-memory}, VRGNet \cite{wang-cvpr21-from} are selected to make comparisons. PSNR and SSIM \cite{wang-tip04-image} are selected to assess different works quantitatively.

\begin{figure}[t]
\centering
\begin{minipage}{0.7\linewidth}
\centering{\includegraphics[width=1\linewidth]{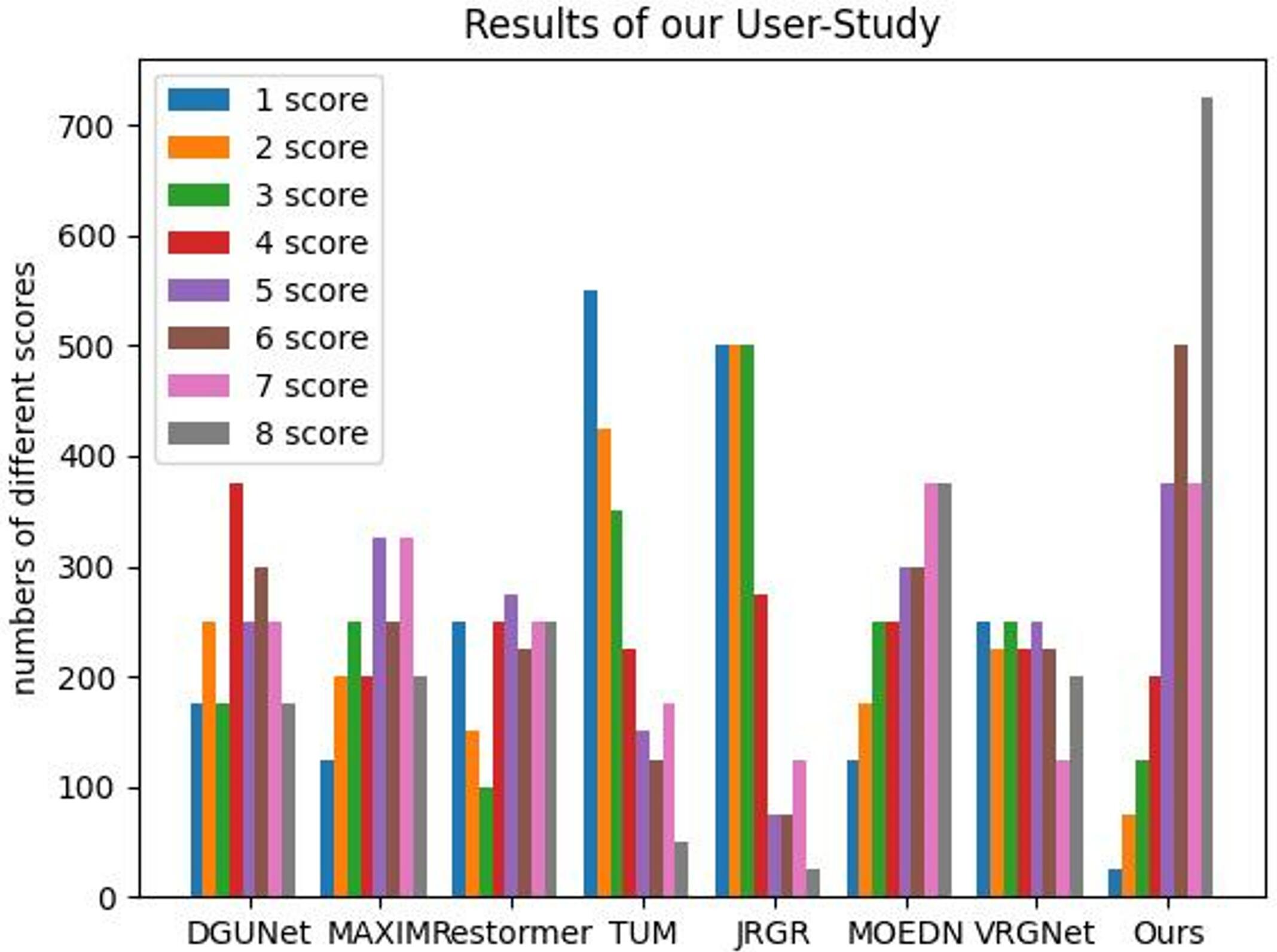}}
\end{minipage}
\caption{Rating statistics of different methods in user-study.}
\label{fig:user-study}
\end{figure}

\subsection{Comparison with State-of-the-Arts}
\label{sec:expe_sota}

\noindent\textbf{Quantitative evaluation on synthetic data.}~~Table \ref{tab:psnr_ssim_sota} reports PSNR and SSIM of different methods on the three synthetic testing datasets. We observe that the semi-supervised methods, i.e., JRGR, MOEDN, VRGNet, are inferior in quantitative evaluation on synthetic datasets. Supervised methods, i.e., DGUNet, MAXIM, Restormer, produce better results than semi-supervised ones. The performance of TUM which handles multiple tasks with single network is relatively lower. This may be due to the fact that the deep representations for different tasks are disparate, such diversity can harm the performance of each other. Our method achieves the best performance on all of these datasets. This is because our proposed task transfer strategy learns highly effective deraining deep representation.


\begin{figure}[t]
	\begin{center}
		\begin{minipage}{0.24\linewidth}
			\centering{\includegraphics[width=1\linewidth]{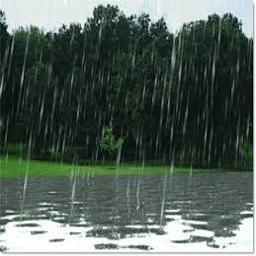}}
			\centerline{Input}
		\end{minipage}
		\hfill
		\begin{minipage}{0.24\linewidth}
			\centering{\includegraphics[width=1\linewidth]{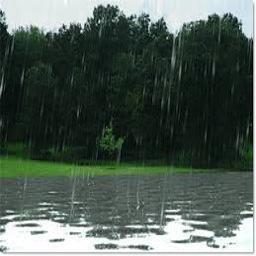}}
			\centerline{w/ $\mathcal{T}_{recog}$}
		\end{minipage}
		\hfill
  \begin{minipage}{0.24\linewidth}
			\centering{\includegraphics[width=1\linewidth]{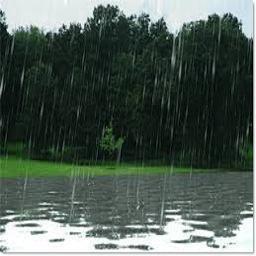}}
			\centerline{w/ $\mathcal{T}_{recon}$}
		\end{minipage}
		\hfill
		\begin{minipage}{0.24\linewidth}
			\centering{\includegraphics[width=1\linewidth]{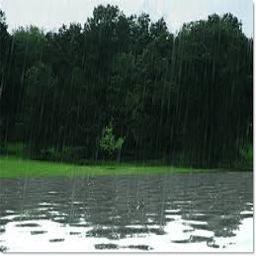}}
			\centerline{Both}
		\end{minipage}
	\end{center}
	\caption{Visual results of ablating $\mathcal{T}_{recog}$ and $\mathcal{T}_{recon}$.}
	\label{fig:visul_comp_sole}
\end{figure}

\noindent\textbf{Qualitative evaluation on synthetic data.}~~Figure \ref{fig:synthetic_compare} visually shows the deraining results of different methods on a synthetic rainy image from Test-III, in which the images resemble real-world ones, so that can better test the generalization of deraining models \cite{wang-cvpr19-spatial}. We observe that the semi-supervised MOEDN, VRGNet and our task transfer method produce better visual quality than the supervised ones, which further illustrates the importance of using real data to improve generalization.

\noindent\textbf{Qualitative evaluation on real data.}~~Real-world rain streaks usually possess blurry edges whose intensities are very close to these of background contents, shown in Figure \ref{fig:practical_compare}, which makes deraining task highly challenging. We observe that the supervised DGUNet, MAXIM, Restormer and the semi-supervised MOEDN produce relatively better visual quality. Another two semi-supervised methods, i.e., JRGR and VRGNet, do not produce satisfactory results, which may be because the pseudo labels produced by these methods for real rainy images are inaccurate. By contrast, our method produces better visual quality, removing more rain streaks and preserving more image details.

\begin{table}[t]
\begin{center}
\begin{tabular}{c|c|c|c}
     & w/ $\mathcal{T}_{recog}$ & w/ $\mathcal{T}_{recon}$ & Both \\
\toprule
    Test-I &  34.34/0.910 & 33.92/0.913 & \textbf{35.01}/\textbf{0.941} \\ 
    Test-II  &  32.61/0.915 & 32.27/0.926 & \textbf{33.23}/\textbf{0.938} \\ 
    Test-III  & 36.58/0.954 & 36.12/0.941 & \textbf{36.65}/\textbf{0.960} \\
\bottomrule
\end{tabular}
\end{center}
\caption{PSNR/SSIM results of ablating $\mathcal{T}_{recog}$ and $\mathcal{T}_{recon}$.}
\label{tab:psnr_ssim_sole}
\end{table}

\noindent\textbf{User study on real data.}~~To study the generalization of different methods completely, we invite 20 colleagues to rank the results of $8$ methods on $100$ real rainy images in terms of removing more rain streaks and preserving more details. The best deraining result of an image is scored $8$, the second best is scored $7$, etc. The rating results of the user study are given in Figure \ref{fig:user-study}, which indicates that our method obtains most $8$ score. The total scores of different methods are shown in Table \ref{tab:user-study}. Both Figure \ref{fig:user-study} and Table \ref{tab:user-study} verify that our method receives more approvals from peer workers.

\begin{figure}[t]
	\begin{center}
		\begin{minipage}{0.32\linewidth}
			\centering{\includegraphics[width=1\linewidth]{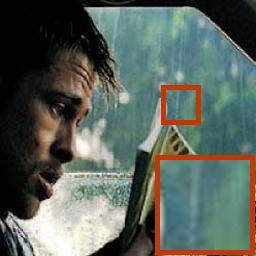}}
			\centerline{(a) Input}
		\end{minipage}
		\hfill
		\begin{minipage}{0.32\linewidth}
			\centering{\includegraphics[width=1\linewidth]{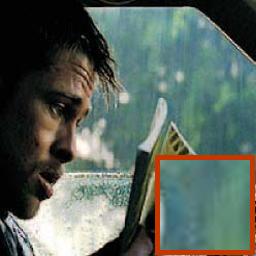}}
			\centerline{(b) $V_1$}
		\end{minipage}
		\hfill
		\begin{minipage}{0.32\linewidth}
			\centering{\includegraphics[width=1\linewidth]{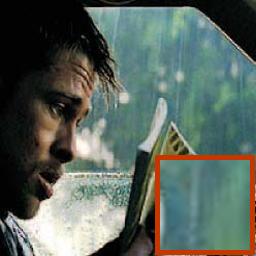}}
			\centerline{(c) $V_2$}
		\end{minipage}
	\end{center}
	\caption{Visual results of $V_1$ and $V_2$ on real-world rainy images. (a) Input. (b) Results of $V_1$. (c) Results of $V_2$.}
	\label{fig:differ_decode}
\end{figure}

\subsection{Ablation Study}
\label{sec:disc}

\begin{table}[t]
\begin{center}
\begin{tabular}{c|c|c}
		 & $V_1$ & $V_2$  \\
\toprule
		Test-I & \textbf{35.01}/\textbf{0.941}  & 34.26/0.927 \\ 
	    Test-II  & \textbf{33.23}/\textbf{0.938} & 32.75/0.931 \\ 
		Test-III  & \textbf{36.65}/\textbf{0.960} & 35.79/0.944 \\
		\hline
\end{tabular}
\end{center}
\caption{PSNR/SSIM of two training strategies.}
\label{tab:psnr_ssim_differ_decode}
\end{table}

\noindent\textbf{Ablation on transfer tasks.}~~In our work, we design two transfer tasks, i.e., $\mathcal{T}_{recog}$ and $\mathcal{T}_{recon}$, to learn effective prior representation for our target task $\mathcal{T}_{dr}$. Here, we study their separate effects on $\mathcal{T}_{dr}$. The results in Table \ref{tab:psnr_ssim_sole} show that $\mathcal{T}_{recog}$ is better than $\mathcal{T}_{recon}$ when used separately, and combining them together performs the best. In Figure \ref{fig:visul_comp_sole}, we observe that $\mathcal{T}_{recog}$ removes more rain streaks, while $\mathcal{T}_{recon}$ preserves more image details. Also, their combination obtains better result. This study illustrates that these two transfer tasks learn complementary deraining representations.

\noindent\textbf{Effectiveness of transfer tasks}~~To verify the effectiveness of transfer tasks to learn deraining representation, we denote our proposed learning strategy as $V_1$, and define another strategy $V_2$ where the same structure ($\mathcal{E}_{stu}(\cdot)$ + $\mathcal{D}_{dr}(\cdot)$) as $V_1$ is used, but it is trained from scratch using only synthetic data. Table \ref{tab:psnr_ssim_differ_decode} compares $V_1$ and $V_2$ on the synthetic testing datasets, Test-I, Test-II, and Test-III. It clearly shows that our method $V_1$ outperforms $V_2$. Visual example is given in Figure \ref{fig:differ_decode}, showing that $V_1$ removes more rain streaks.

\section{Conclusion}
\label{sec:conc}
In this paper, we proposed a new learning strategy, i.e., task transfer learning, to improve generalization of deraining models by using real data. We first discovered the substantial reason for the low generalization of supervised deraining models to real rainy scenes by detailed statistic analysis. Inspired by our studies, we design two simple yet highly effective transfer tasks, i.e., recognition of rain and rain-free scenes, reconstruction of blurred images, to learn deraining-favorable representation from real data. Our method beats state-of-the-art approaches on both synthetic and real data. Moreover, we build a new accurate real-world rainy dataset purely via capturing rainy scenes with a camera, which ensures the effectiveness of the collected rainy data and facilitates further generalization researches of deraining models.


\end{document}